\documentclass[conference]{IEEEtran}
\IEEEoverridecommandlockouts
\usepackage{cite}
\usepackage{amsmath,amsthm,amssymb,amsfonts}
\usepackage{graphicx}
\usepackage{textcomp}
\usepackage{xcolor}

\def\BibTeX{{\rm B\kern-.05em{\sc i\kern-.025em b}\kern-.08em
    T\kern-.1667em\lower.7ex\hbox{E}\kern-.125emX}}

\title{Spatiotemporally adaptive compression for scientific dataset with feature preservation -- a case study on simulation data with extreme climate events analysis 
}
\author{\IEEEauthorblockN{
Qian Gong\IEEEauthorrefmark{1},
Chengzhu Zhang\IEEEauthorrefmark{2}, 
Xin Liang\IEEEauthorrefmark{3}, 
Viktor Reshniak\IEEEauthorrefmark{1}, 
Jieyang Chen\IEEEauthorrefmark{4}, 
Anand Rangarajan\IEEEauthorrefmark{5}, 
Sanjay \\ Ranka\IEEEauthorrefmark{5},
Nicolas Vidal\IEEEauthorrefmark{1}, 
Lipeng Wan\IEEEauthorrefmark{6}, 
Paul Ullrich\IEEEauthorrefmark{2}, 
Norbert Podhorszki\IEEEauthorrefmark{1}, 
Robert Jacob\IEEEauthorrefmark{7},
Scott Klasky\IEEEauthorrefmark{1}
}
\IEEEauthorblockA{{\textit{Oak Ridge National Laboratory}, Oak Ridge, TN}\IEEEauthorrefmark{1} \\
{\textit{Lawrence Livermore National Laboratory}, Livermore, CA}\IEEEauthorrefmark{2} \\
{\textit{University of Kentucky}, Lexington, KY}\IEEEauthorrefmark{3} \\
{\textit{University of Alabama}, Birmingham, AL}\IEEEauthorrefmark{4} \\
{\textit{University of Florida}, Gainesville, FL}\IEEEauthorrefmark{5}\\
{\textit{Georgia State University}, Atlanta, GA}\IEEEauthorrefmark{6}\\
{\textit{Argonne National Laboratory}, Lemont, IL}\IEEEauthorrefmark{7}\\
Email: \IEEEauthorrefmark{1}gongq@ornl.gov}
}

\usepackage[T1]{fontenc}
\usepackage{graphicx}
\usepackage{booktabs}
\usepackage{longtable}
\usepackage{pbox}
\usepackage{mathtools}
\usepackage{caption}
\usepackage{subcaption}
\usepackage{algpseudocode}
\usepackage{algorithm}
\usepackage{siunitx}
\usepackage[inline]{enumitem}
\usepackage{siunitx}
\usepackage{diagbox}
\usepackage[]{mdframed}

\usepackage{comment}
\usepackage[normalem]{ulem}

\DeclarePairedDelimiter\abs{\lvert}{\rvert}

\theoremstyle{plain}

\usepackage{xparse}
\NewDocumentCommand\N{sm}{\mathcal{N}\IfBooleanT#1{^{\ast}}_{#2}}

\makeatletter
\NewDocumentCommand{\@Coefficients}{m}{\text{\ttfamily\upshape #1}}
\newcommand\uMultilevelCoefficients{\@Coefficients{u\char`_mc}}
\newcommand\newMultilevelCoefficients{\@Coefficients{\~{u}\char`_mc}}
\makeatother

\NewDocumentCommand\at{m}{\text{\upshape\ttfamily\lbrack}#1\text{\upshape\ttfamily\rbrack}}

\begin{document}

\maketitle

\begin{abstract}
Scientific discoveries are increasingly constrained by limited storage space and I/O capacities. For time-series simulations and experiments, their data often need to be decimated over timesteps to accommodate storage and I/O limitations. In this paper, we propose a technique that addresses storage costs while improving post-analysis accuracy through spatiotemporal adaptive, error-controlled lossy compression. We investigate the trade-off between data precision and temporal output rates, revealing that reducing data precision and increasing timestep frequency lead to more accurate analysis outcomes. Additionally, we integrate spatiotemporal feature detection with data compression and demonstrate that performing adaptive error-bounded compression in higher dimensional space enables greater compression ratios, leveraging the error propagation theory of a transformation-based compressor.

To evaluate our approach, we conduct experiments using the well-known E3SM climate simulation code and apply our method to compress variables used for cyclone tracking. Our results show a significant reduction in storage size while enhancing the quality of cyclone tracking analysis, both quantitatively and qualitatively, in comparison to the prevalent timestep decimation approach. Compared to three state-of-the-art lossy compressors lacking feature preservation capabilities, our adaptive compression framework improves perfectly matched cases in TC tracking by 26.4-51.3\% at medium compression ratios and by 77.3-571.1\% at large compression ratios, with a merely 5-11\% computational overhead.
    
\end{abstract}


\section{Introduction}

In the realm of scientific simulations and experiments, the exponential growth of computational power has led to a significant increase in data generation. The limited capacities of storage and I/O systems have now become a bottleneck, hindering the progress of scientific discoveries and analyses. 
For instance, the Oak Ridge Leadership Computing Facility recently reported that their new supercomputer Frontier~\cite{frontier} delivers $8\times$ the peak performance of its predecessor Summit~\cite{summit}, while storage and I/O improvements remains modest at $2-4\times$. 
To address the challenges associated with storage and query efficiency, data compression has emerged as an effective solution~\cite{chang2004retrieving, deri2012tsdb,chen2001query, arion2007xquec}. However, conventional lossless compression methods often perform poorly on scientific data due to the random mantissa in floating-point formats, with a compression ratio less than 2 in most cases. Timestep decimation is a prevalent reduction approach used in scientific simulations and experiments that require recording time-series data over extended periods. However, numerous studies \cite{zarzycki2017assessing, zarzycki2021metrics, mcclenny2020sensitivity, zhou2021uncertainties} have shown that higher temporal resolutions yield more accurate results for process-oriented and phenomena-based model evaluations. Outputting model data at decimated timesteps risks degrading the fidelity of downstream scientific analysis.

To overcome these challenges, error-controlled lossy compressors have emerged as an alternative approach for scientific applications that require significant data reduction rates while preserving accuracy \cite{tao2017significantly, liang2018error, zhao2021optimizing, lindstrom2006fast, lindstrom2014fixed, ainsworth2017compression, ainsworth2019multilevel, ainsworth2019qoi}. However, most existing compressors lack information-awareness, making them suboptimal for tasks involving the preservation of quantities of interest (QoI). QoIs refer to physical properties and typologies that are the targets of post-hoc analysis. While recently a few papers have proposed methods to control derived errors during data compression \cite{ainsworth2019multilevel, jiao2022toward, liang2020toward}, they either focus on specific features or tend to overestimate the required error bounds, resulting in suboptimal compression ratios.

In this paper, we compare the two data reduction techniques, lossy compression and temporal decimation, and investigate their respective impacts on post-analysis outcomes. Additionally, we propose a novel approach that combines data reduction with two techniques: trading precision for higher temporal resolution and spatiotemporal feature tracking. Our goal is to provide a general error-bounded compression framework that significantly reduces data size while preserving complex QoIs derived from spatiotemporal analyses.

Designing an efficient compressor for spatiotemporal QoI preservation poses several challenges. First, it is crucial to understand the trade-off between spatial precision used for compression and the temporal frequency adopted for decimation, and their respective impacts on QoIs. Second, since the types and locations of QoIs are usually unknown prior to data compression, designing a universal method for preserving multiple QoIs is challenging. Finally, enabling QoI preservation often incurs high overhead, as identifying and storing QoI-related regions requires additional computational and storage costs. To address these challenges, we propose a set of methods that efficiently and effectively tackle these issues.

The contributions of our work are summarized as follows:
\begin{itemize}
    \item We investigate how timestep decimation impacts the quality and performance of spatiotemporal analyses. Our findings reveal that, at the same storage cost, data saved at reduced precision but higher temporal frequencies can better preserve spatiotemporal QoIs than uncompressed data outputted at decimated timesteps.
    \item We redesign a critical region detection method proposed in a previous work \cite{gong2022region} to provide more accurate spatiotemporal feature tracking. Unlike other QoI-preserving compressors \cite{liang2022toward, jiao2022toward}, which tie to specific analyses, our method can be applied to a wide range of QoIs involving spatiotemporal features.
    \item We extend pointwise error preservation theories of a transformation-based compressor to high-dimensional space, which is crucial for enabling spatiotemporal feature-oriented compression.
    \item We validate our method with a mission-critical climate QoI, TC (Tropical Cyclone) tracking, using data generated from a real-world application, Energy Exascale Earth System Model (E3SM) \cite{caldwell2019doe}. TC is an impactful weather phenomenon that can cause significant loss of life and economic damage \cite{mendelsohn2012impact}. Our experiments demonstrate that the proposed method provides higher accuracy for TC tracking while achieving higher compression ratios at a small computational overhead compared to both timestep decimation and uniform compression.
\end{itemize}

The rest of the paper is organized as follows. Section~\ref{sec:related} discusses related works and provides technical background. Section~\ref{sec:compression} introduces our proposed methods, including trading precision for temporal resolution, region-wise error control, and critical region detection. Section~\ref{sec:evaluation} presents the evaluation results on TC tracking using E3SM data. Finally, Section~\ref{sec:conclusion} concludes the paper and outlines future research directions.

\section{Problem and Research Background}\label{sec:related}
\subsection{Scientific data compression}
\label{sec:bkg-lossyCompression}
Compressors can be categorized into lossless and lossy compression methods. Lossless compressors, such as GZIP~\cite{deutsch1996gzip}, FPC~\cite{burtscher2008fpc}, and ZSTD~\cite{collet2021rfc}, rely on encoding techniques to transform data into compact representations. However, their compression ratios are often limited when applied to floating-point scientific data.
In contrast, lossy compression methods can achieve higher compression ratios by sacrificing accuracy. The ability to control errors is crucial for the practical application of lossy compression in scientific domains. 
State-of-the-art lossy compressors in this field include ISABELA~\cite{lakshminarasimhan2013isabela}, SZ~\cite{tao2017significantly, zhao2021optimizing}, FPZIP~\cite{lindstrom2006fast}, ZFP~\cite{lindstrom2014fixed}, and MGARD~\cite{ainsworth2018multilevel, ainsworth2019multilevel, ainsworth2019qoi}.
Compared to lossless methods, lossy compression typically involves additional steps, such as data decorrelation and quantization, which transforms floating-point data into coefficients that are amenable to compression then to a discrete set with lower entropy, and mathematical theories to ensure user-prescribed error bounds. 

A recent trend in scientific data compression is the preservation of accuracy for Quantities of Interest (QoIs) that are crucial for post-hoc scientific analysis. Existing approaches achieve this by deriving either a global error bound for all data or local error bounds for individual data points or regions. MGARD~\cite{ainsworth2019qoi} and constraint satisfaction-based post-processing~\cite{lee2022error} are representative approaches in the former category. These methods demonstrate significant compression ratios on QoIs such as density or weighted sum, which are integral over extended space~\cite{gong2021maintaining,banerjee2022algorithmic}. However, the use of a uniform global error bound can lead to over-preservation when features are local. Furthermore, the theories underlying the above approaches rely on linear theory, making their generalization to preserve complex non-linear QoIs challenging.
Liang et al.\cite{liang2020toward} and Pu et al.\cite{jiao2022toward} propose localized QoI preservation methods based on a prediction-based compressor, SZ. Their approaches support error control for four QoI classes, including polynomials, logarithmic mapping, weighted sum, and critical point/isosurfaces. The extension to support more categories and complex QoIs, such as TC tracking, are challenging as these methods need to be specifically designed to establish the mapping between errors in the original data and errors in the QoIs. Additionally, they suffer from high computational overhead (2-5 times slower on compression and up to 10 times slower on decompression compared to standard SZ) and limited compression ratios due to the requirement of compressing and storing the derived local error bounds. 

\subsection{MGARD lossy compression model}
\label{sec:mgard-decomposition}
Our spatiotemporal adaptive compression framework is based on the utilization of MGARD, which is an error-controlled lossy compressor employing multilinear interpolation in data decomposition. This approach allows us to track variational features by leveraging the residuals of interpolation without incurring additional computations~\cite{gong2022region}. MGARD makes use of multigrid theories. 
Given an input array \texttt{u} in \(d\)-dimensions defined on a grid $\N{L}$, MGARD applies multilinear interpolation and $L^2$ projection to decompose the data. This transformation yields a set of multilevel coefficients, denoted as $\uMultilevelCoefficients$, residing on subgrids $\N{L - 1}, \dotsc, \N{l}, \dotsc, \N{0}$, downsampled from $\N{L}$. After decomposition, MGARD applies linear-scaling quantization, Huffman encoding~\cite{moffat2019huffman}, and GZIP/ZSTD compression to generate a reduced representation of $\uMultilevelCoefficients$, denoted as \(\newMultilevelCoefficients\). 
The reconstruction process recomposes \(\newMultilevelCoefficients\) back to \(\tilde{u}\) defined on the original grid $\N{L}$. 
The difference between \(\uMultilevelCoefficients\) and the quantized \(\newMultilevelCoefficients\) induces a compression error between \(u\) and \(\tilde{u}\). 
MGARD mathematically ensures the errors adheres to user-prescribed bounds defined in 
\(L^{2}\), \(L^{\infty}\), or $H^s$ norms, where $H^s$ is the Sobolev space with an index $s$ \cite{ainsworth2019multilevel,ainsworth2018multilevel,ainsworth2019qoi}. 

Unlike predictor-based compressors~\cite{tao2017significantly, zhao2021optimizing}, where quantization errors can be maintained on a point-by-point basis, in MGARD, the compression error at a point $y$ is determined by recomposing and accumulating the quantization errors projected from the multilevel coefficients at $y$ and the surrounding locations. 

For a node \(x\) at level $l$ in uniform grid space, the work in \cite{gong2022region} prove that the compression error at the node $y$, induced by quantizing the multilevel coefficient $\uMultilevelCoefficients\at{x}$, can be expressed as follows:
\begin{equation}
\abs{(\texttt{u} - \tilde{\texttt{u}})(y)}
\leq C_d \sum_{l = 0}^{L} \sum_{x \in \N*{l}} \abs[\big]{\Delta\uMultilevelCoefficients\at{x}} 
(2+\sqrt{3})^{-d_{l-1}(x, y)},
\label{eq:point-wise-error}
\end{equation}
where \(\Delta\uMultilevelCoefficients\at{x} = \uMultilevelCoefficients\at{x} - \newMultilevelCoefficients\at{x}\), $C_d$ is a scale factor relevant to the data dimension, and \(d_{l-1}(x, y) = \abs{x - y} / h_{l-1}\), which represents the number of grid spacing between $x$ and $y$ at the next coarser level. 

Equation~\ref{eq:point-wise-error} demonstrates that the point-wise quantization error follow an oscillatory function that decays exponentially with the distance from \(x\) after recomposition. 
This exponential decay rate reveals that the compression error in the region of interest (RoI) is primarily determined by the quantization of the coefficients within the region itself and in a set of narrow surrounding zones across the sub-grids. To prevent the propagation of compression errors between regions, previous research~\cite{gong2022region} suggested to construct a buffer zone around the RoIs and compress it using the same compression error bound as RoIs. This operation ensures that a compression error of $\tau_1$ at a node outside the RoIs will be reduced to approximately $\tau_1\times C_d(2+\sqrt{3})^{-R_{bz}}$ when it propagates to the edge of RoIs. 

\section{Spatiotemporal Data Compression}\label{sec:compression}
This paper focuses on spatiotemporal field data generated by simulations in a per-timestep basis, where the QoIs require variational feature tracking in temporal domain. To address the storage cost and compression incurred errors in post-analysis, we propose three techniques enabling spatiotemporal adaptive lossy compression.

\subsection{Trade data precision for temporal resolution}
\label{sec:trade-off-precision-freq}
For models aiming to capture physical phenomena across a wide range of time scales, timestep outputs often needs to be recorded. Due to limited storage and I/O capacity, storing simulation data at their physical discretization is often infeasible. While precision-based lossy compression methods have shown promise in reducing the storage cost~\cite{cappello2019use, gong2021maintaining, diffenderfer2019error}, the prevalent approach used by scientific models for data reduction is through timestep decimation. This is because many application scientists believe that the original time-series data recorded at lower rates are more "trustable" than precision-reduced data recorded with higher temporal frequency. Therefore, the first key question we explore in this paper is: between the reduction in precision (e.g., through lossy compression) and the reduction in temporal resolution (e.g., through timestep decimation), which method has a greater impact on the accuracy of spatiotemporal analyses?

\begin{figure}[t]
  \centering
  \begin{subfigure}[b]{0.475\linewidth}
    \centering
     \includegraphics[width=\linewidth]{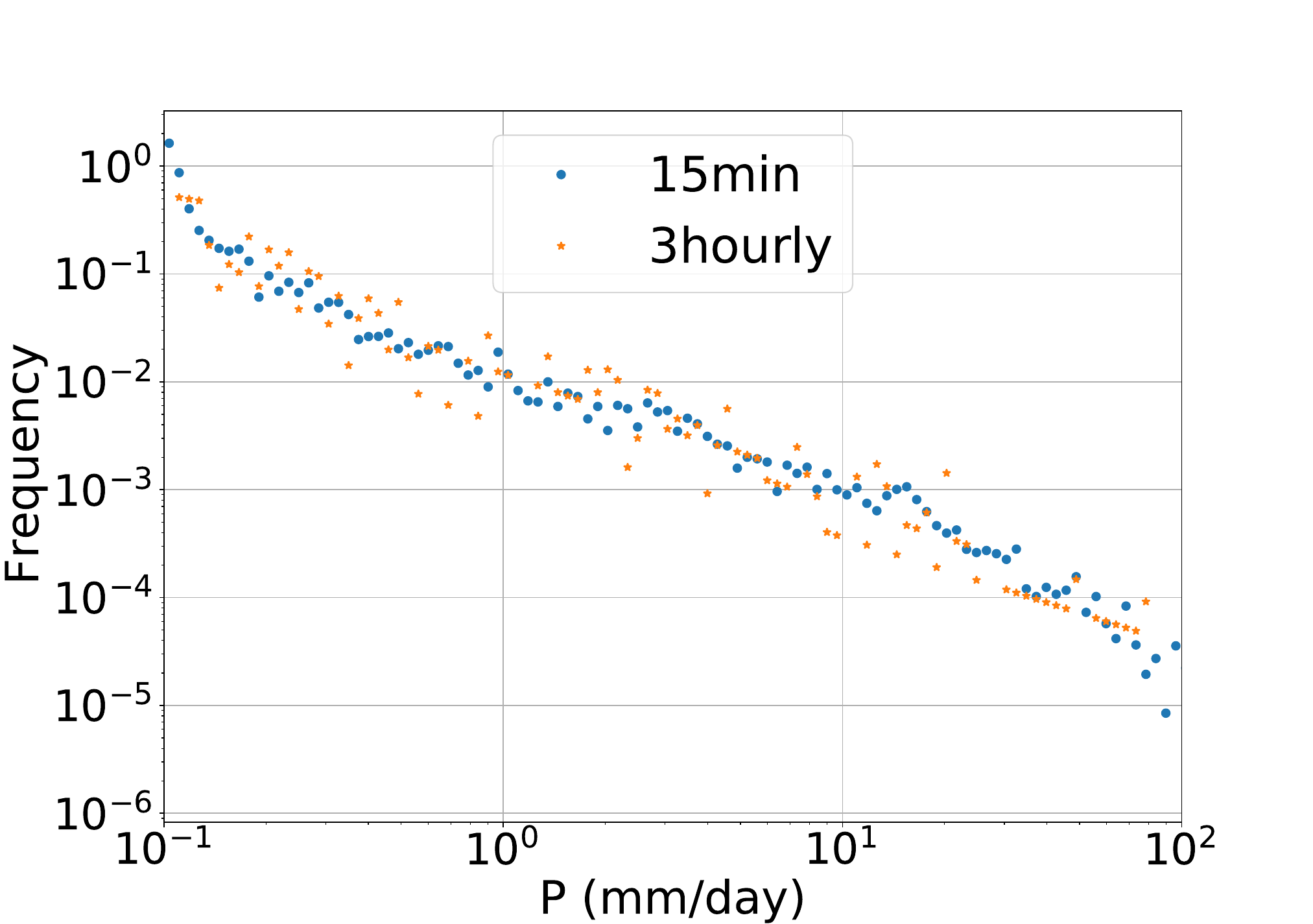}
     \caption{Oklahoma}
     \label{fig:precip_freq_SGP}   
  \end{subfigure}
\hfill
\begin{subfigure}[b]{0.475\linewidth}
  \centering
  \includegraphics[width=\linewidth]{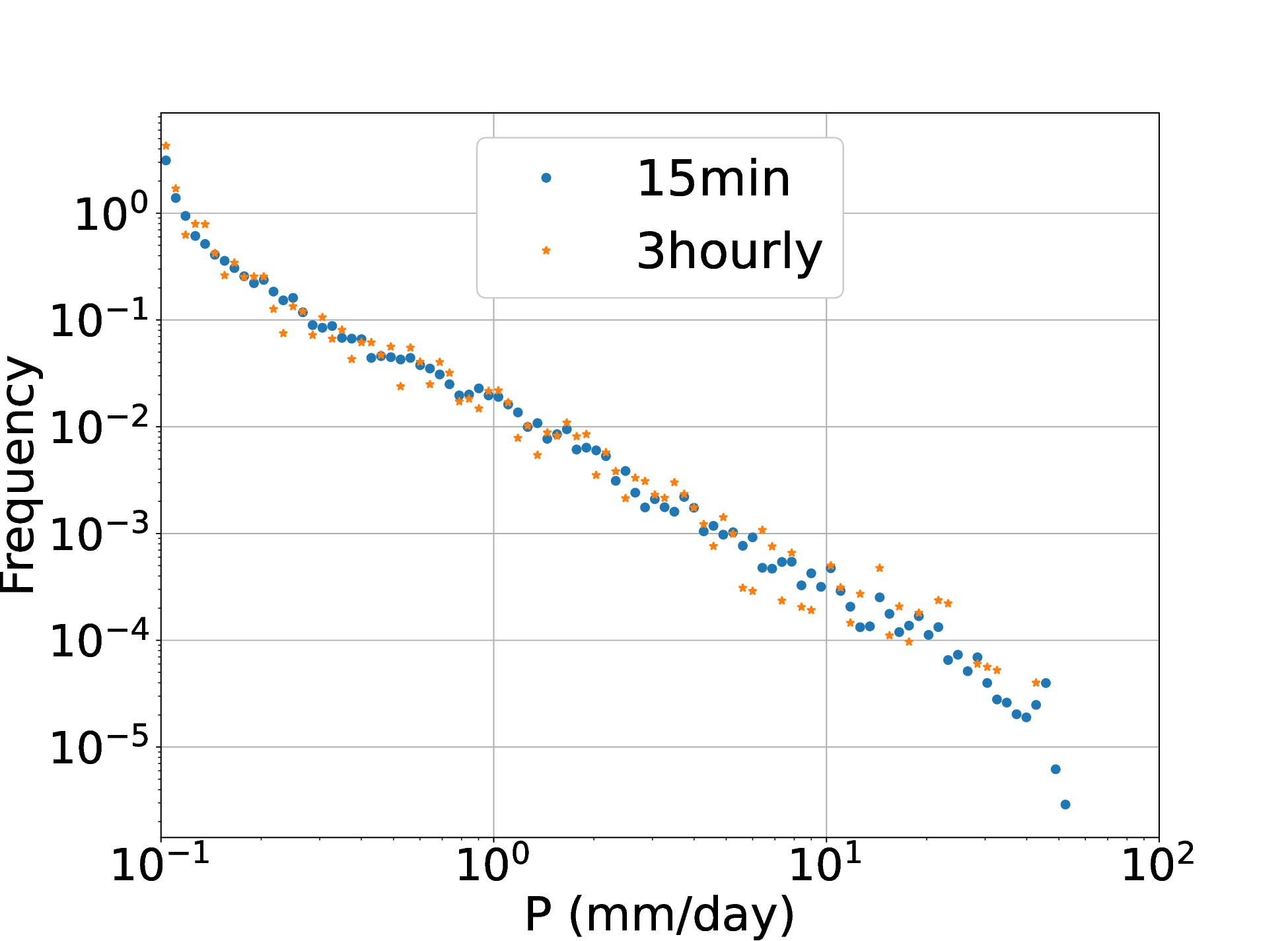}
  \caption{Alaska}
  \label{fig:precip_freq_NSA}
  \end{subfigure}
  \caption{Frequency distributions of precipitation intensity over two selected grid points within (a) Oklahoma and (b) Alaska using 15-minute and 3-hourly E3SM output.}     
\end{figure}

In contrast to downstream QoIs which can be explicitly defined (e.g., kinetic energy $E=\frac{1}{2}mv^2$), the detection of spatiotemporal events, such as weather phenomena, is influenced by the choice of detection algorithm. Analyzing spatiotemporal data typically involves the instantaneous "extraction" of features at each snapshot and spatiotemporal "tracing" to merge and track the featured event propagation. In this study, we use "detection" and "tracking" interchangeably to describe the procedure of feature extraction and spatiotemporal tracing. Temporal decimation can affect detection by potentially missing short-lived phenomena and introducing discontinuities in event tracing. On the other hand, reduction in precision preserves temporal connectivity but alters the detection results by introducing shifted or false-appeared/disappeared topological features. Our findings suggest that, for QoIs involving the tracing of event temporal propagation, reducing precision offers several potential benefits compared to temporal decimation.

\begin{mdframed}[skipabove=2mm]
\textbf{Finding 1: High temporal resolution is crucial for short-lived extremes detection.}
\end{mdframed}
\vspace{2mm}

To illustrate this point, we consider short-duration events in climate and weather phenomena. Flash flood induced by intensive precipitation may last only a few hours with extreme intensities captured at sub-hourly intervals~\cite{berg2018precipitation, meyer2021more}. However, climate simulation codes such as E3SM often makes model output at 6-hourly and 3-hourly rate instead the physical timestep, which is 15-minute.
Figure~\ref{fig:precip_freq_SGP} and \ref{fig:precip_freq_NSA} show the simulated frequency distribution of instantaneous precipitation rates at 15-minute intervals and 3-hourly intervals over two point locations in Oklahoma, USA, and Alaska, USA. The extreme intensities are clearly missed in the 3-hourly output. As climate models move towards higher spatial and temporal resolutions to capture more detailed processes, it becomes increasingly critical to save data at the physical temporal resolution of the simulation. 

\begin{mdframed}[skipabove=2mm]
\textbf{Finding 2: Spatiotemporal analyses conducted on precision-reduced data exhibit greater resilience to outliers caused by the design choice of analysis algorithms compared to temporally decimated data.}
\end{mdframed}
\vspace{2mm}

\begin{figure}[t]
  \centering
     \includegraphics[width=0.7\linewidth]{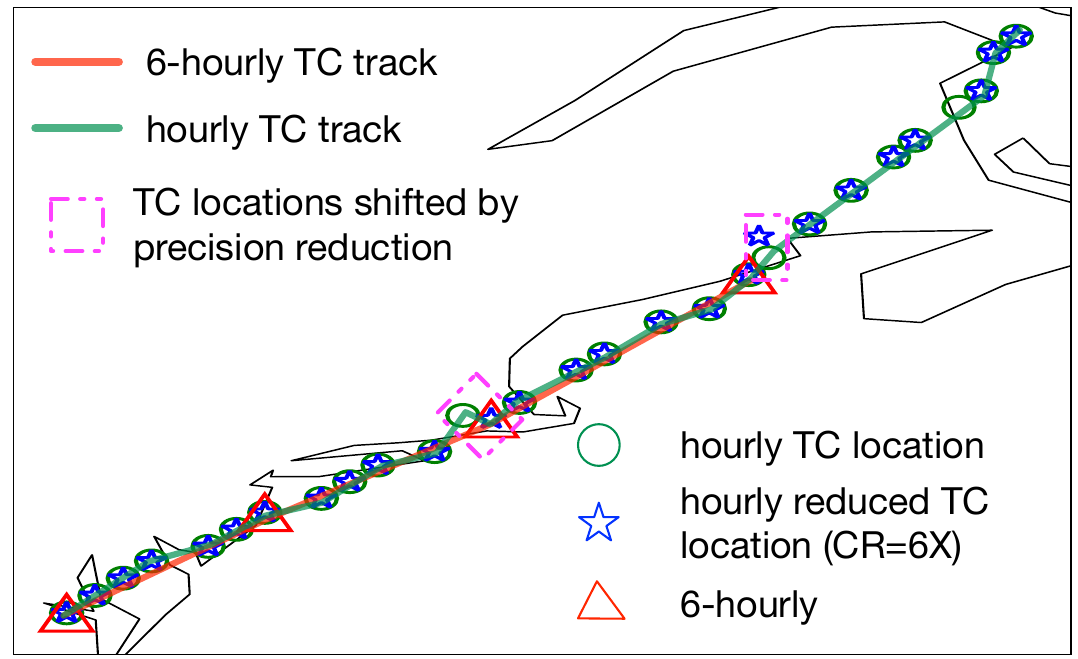}
     \caption{Example Tropical Cyclone (TC) tracks detected in the hourly, hourly lossy compressed, and temporally decimated (6-hourly) data. With a higher temporal resolution, lossy compression causes less errors on the detected TC trajectory than data reduced by temporal decimation.}
     \label{fig:benefit_hifreq}   
\end{figure}

We use TC tracking as an example to further illustrate the benefits of trading precision for higher temporal resolution. A TC track is formed by detecting footprints across consecutive time steps and connecting candidates that satisfy certain criteria. 
Figure~\ref{fig:benefit_hifreq} demonstrates the TC cyclone footprints and the stitched track detected using the original hourly dataset, the hourly dataset compressed by a factor of 6 using MGARD, and the 6-hourly temporally decimated dataset outputted from the same simulation run. As the figure illustrates, the track obtained from the 6-hourly dataset is significantly shorter than the original measurement. The reduced length of the TC track produced by the 6-hourly data is caused by a missing candidate in the last detection. 
In comparison, a few shifted cyclone footprints found in hourly compressed dataset do not change the cyclone trajectory. The TC track detected from the hourly uncompressed and lossy compressed datasets are almost identical.

\begin{mdframed}[skipabove=2mm]
\textbf{Finding 3: Temporal redundancies in time-series data recorded at frequent rates contribute to larger compression ratios.}
\end{mdframed}
\vspace{2mm}

\begin{figure}[t]
  \centering
     \includegraphics[width=0.9\linewidth]{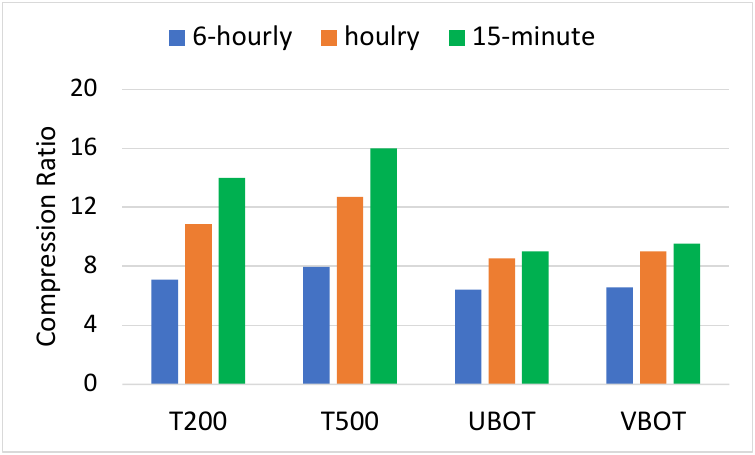}
     \caption{Compression ratios achieved on 4 variables outputted from E3SM simulation at every 15-minute, hourly, and 6 hourly rate when buffer 120 timesteps of data in memory and compress them at once, using a relative RMSE of \num{1e-3}.}
     \label{fig:cr_2d_3d}   
\end{figure}

Buffering and compressing time-series data as a whole can leverage the temporal correlations, resulting in higher data compression ratios. Despite the finer temporal discretization leading to larger data volumes, the increased correlation among timestep data enhances the compressbility of the data. To demonstrate this, we consider four variables obtained from an E3SM simulation run, recorded at 15-minute, hourly, and 6-hourly intervals. For each output rate, we buffer 120 consecutive timesteps in memory and then compress the them together using MGARD with a uniform relative error bound of \num{1e-3} under the $L^2$ norm. Figure~\ref{fig:cr_2d_3d} illustrates that the temporally buffered hourly outputs achieve approximately $33\%$ to $60\%$ larger compression ratios among different variables compared to the 6-hourly outputs. Furthermore, the 15-minute outputs exhibit an additional $5\%$ to $29\%$ improvement compared to the hourly outputs, emphasizing the impact of temporal correlation on compression efficiency.

\subsection{Non-uniform compression for spatiotemporal data}
\label{sec:error-control}
Since information is often non-uniformly distributed in time-space, using different error bounds for data inside and outside of the RoIs helps to achieve both large compression ratios and accurate preservation of QoIs. However, the multilevel decomposition of MGARD introduces a challenge where quantization errors at a single point propagate throughout the entire data space. According to the error propagation and the derived buffer zone theories in Section~\ref{sec:mgard-decomposition}, the error bound used for compressing non-RoI data cannot be arbitrarily large, as it would require a wider buffer zone, which would, in turn, adversely impact the overall compression ratios. Mathematically, the maximally allowed error at a non-RoI data point, $\tau_1$, is expressed as: $\tau_1 < (2+\sqrt{3})^{R_{rz}} \tau_0 / C_d$, where $\tau_0$ is an user-prescribed error bound on data in the RoI, $C_d$ is a scalar, and $R_{rz}$ is the width of buffer zone, i.e., the minimal distance between any RoI and non-RoI data points. 

\begin{figure*}[h]
        \centering
        \begin{subfigure}[b]{0.235\linewidth}
        \centering
        \includegraphics[width=\linewidth]{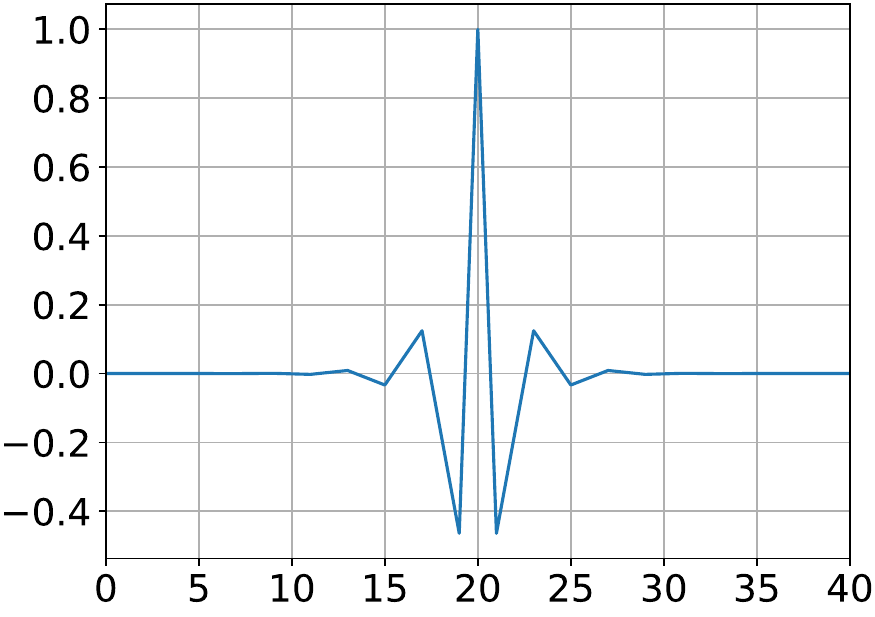}
        \caption{1{\scshape d} space}
        \label{fig:error-prop-1d}
        \end{subfigure}
        \begin{subfigure}[b]{0.235\linewidth}
        \centering
        \includegraphics[width=\linewidth]{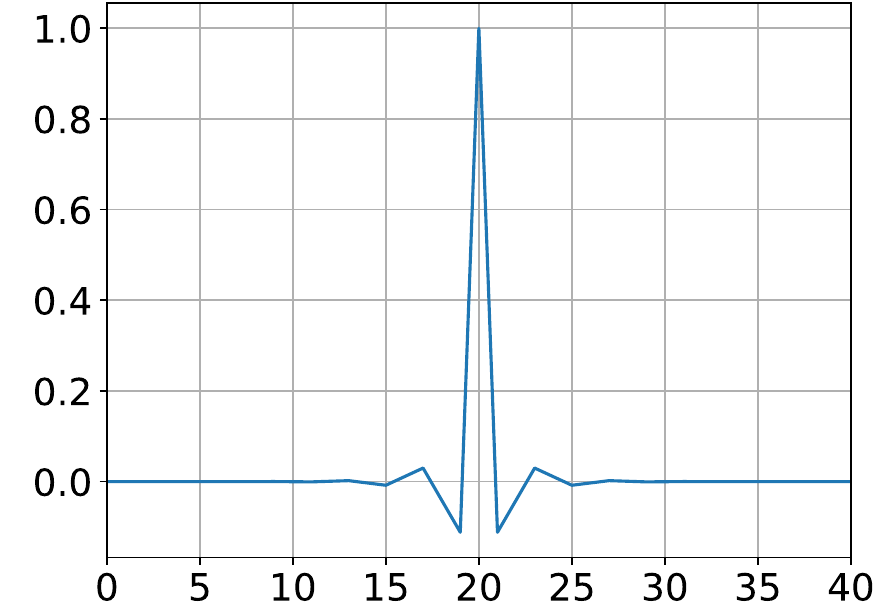}
        \caption{2{\scshape d} space}
        \label{fig:error-prop-2d}
        \end{subfigure}
        \begin{subfigure}[b]{0.235\linewidth}
        \centering
        \includegraphics[width=\linewidth]{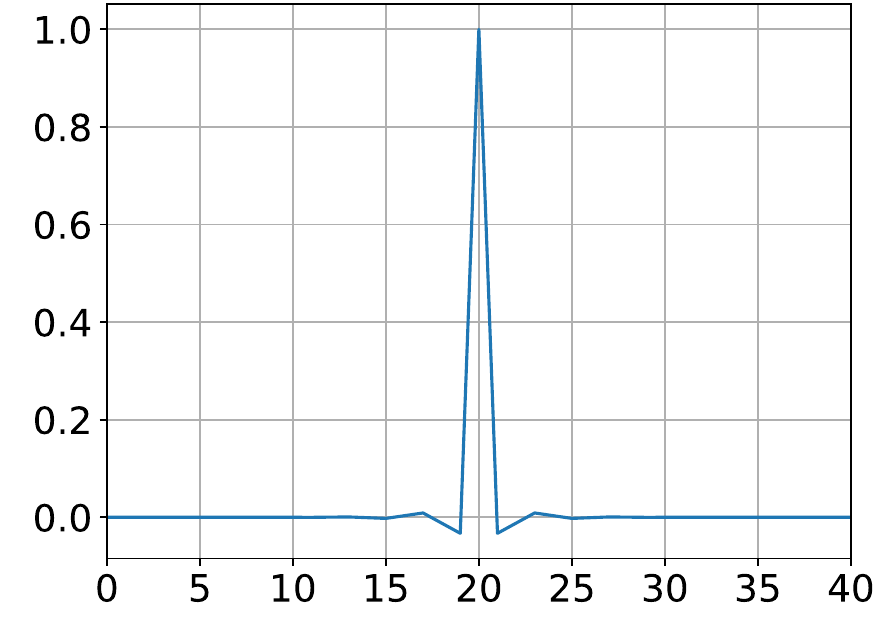}
        \caption{3{\scshape d} space}
        \label{fig:error-prop-3d}
        \end{subfigure}
        \caption{The propagation of compression error induced by a quantization error at a node on a coarser level after the multilevel recomposition in 1{\scshape d}, 2{\scshape d}, and 3{\scshape d} space.}
    \label{fig:error-prop}
\end{figure*}

As mentioned earlier, MGARD decomposition involves a multilinear interpolation and a $L^2$ projection. The computation of the \textit{correction} matrix, $M_l$, used in MGARD's $L^2$ projection, plays a crucial role in determining $C_d$. Mathematically, the multidimensional correction computation can be expressed as a composition of multiple 1{\scshape d} computations along each dimension, i.e.,
$M_l = M_l^1\otimes ... \otimes M_l^d$. 
For instance, in the case of 3{\scshape d} data, the overall correction matrix is obtained by performing a row sweep for row correction, followed by a column sweep on the resulting row correction, and finally a height sweep on the resulting column/row correction. Consequently, the projected values in higher-dimensional space are more suppressed than those in lower-dimensional space after multilevel recomposition. 
Figure~\ref{fig:error-prop} illustrates the projection of point-wise errors after multilevel recomposition for 1{\scshape d}, 2{\scshape d}, and 3{\scshape d} data. The plots confirm that, among all three cases, the error decay follows the same exponential rate (i.e., $1/(2+\sqrt{3})$), however the magnitude of errors at a distance $d$ decreases from 1{\scshape d} to 3{\scshape d}, due to the diminished value of $C_d$ in higher-dimensional space. 
This relationship between the magnitude of the error project and the dimension space of data compression has not been explored in prior studies, leading to the important finding: 

\begin{mdframed}[skipabove=2mm]
\textbf{Finding 4: Regional compression errors decay at a faster rate in higher-dimensional space.}
\end{mdframed}
\vspace{2mm}

According to the previous finding in Section \ref{sec:trade-off-precision-freq}, treating buffered timesteps as an additional dimension help to exploit additional temporal redundancies. \textbf{Finding 4} further emphasizes the benefits of leveraging higher-dimensional space for compressing data which requires regionally changed precision. By considering the spatially and temporally changed feature as variations in the spatiotemporal domain and employing region-adaptive error preservation techniques in the higher-dimensional space, we can achieve even greater compression ratios, as the use of a smaller $C_d$ in the higher-dimensional space enables more aggressive compression of non-RoI data. 

\subsection{Adaptive critical region detection}
\label{sec:adaptive-region-detection}
In the context of applications where RoI information is not available prior to post-analysis, the adaptive compression pipeline requires a quick assessment to identify candidate regions where features may exist. As discussed in Section~\ref{sec:mgard-decomposition}, the multilevel coefficient of node $x_l$ at level $l$ represents the residual of interpolation taken on nodes at a distance $h_l = \pm 2^{L-l}$ from $x_l$, where $L$ is the maximal decomposition level. By excluding the adjustment made by $L^2$ projection, the magnitude of $\uMultilevelCoefficients\at{x}$ at the level $l$ indicates the extent of data variation in a distance of $h_l$ surrounding the node at $x$.  
In this paper, we refer critical regions as areas that exhibit significant spatial variations or temporal instability. Many features of interest in the field of fluid dynamics, geophysics, and atmospheric science can be defined using variational formulations, such as extreme lines/surfaces, trajectories, and vortices. These regions are likely to contain large-value multilevel coefficients after MGARD decomposition. By using multilevel coefficients, which are intermediate outputs during data compression, rather than external metrics such as gradients, for critical region detection avoid introducing additional computational costs.

Previous work by Gong et al.~\cite{gong2022region} proposed to apply an adaptive mesh refinement (AMR)-based approach to locate regions with significant coefficients. However, the recursive refinement employed in their method assesses global-scale variability only at the coarsest partition and performs subsequent evaluations recursively within each selected cell. This approach makes the detection of RoIs sensitive to the choice of the coarsest partition, the recursive selection results in scattered RoI blocks, which are not ideal for presenting integrated areal features.

To address these issues, we introduce a two-step mesh refinement approach consists the following steps:
\begin{enumerate}
\item Global update: We sort the aggregated values from all partition bins and tag those whose aggregated values exceed a user-prescribed global threshold (e.g., the $p^{th}$ percentile). This global update evaluates the regional coefficients' values from a global perspective. 
\item Local update: For cells whose bins are completely missed in the global update, we perform a local update by tagging the bins whose aggregated values are greater than a local threshold prescribed for individual cells. The local update helps retain local features that may have been missed in the global evaluation and preserves the continuity of RoIs.
\end{enumerate}
At the end of each round of block partition, we combine the bins tagged by both the global and local updates, passing them to the next round of refinement until the termination criteria are met.

\begin{figure}[h]
  \centering
  \includegraphics[width=0.9\linewidth]{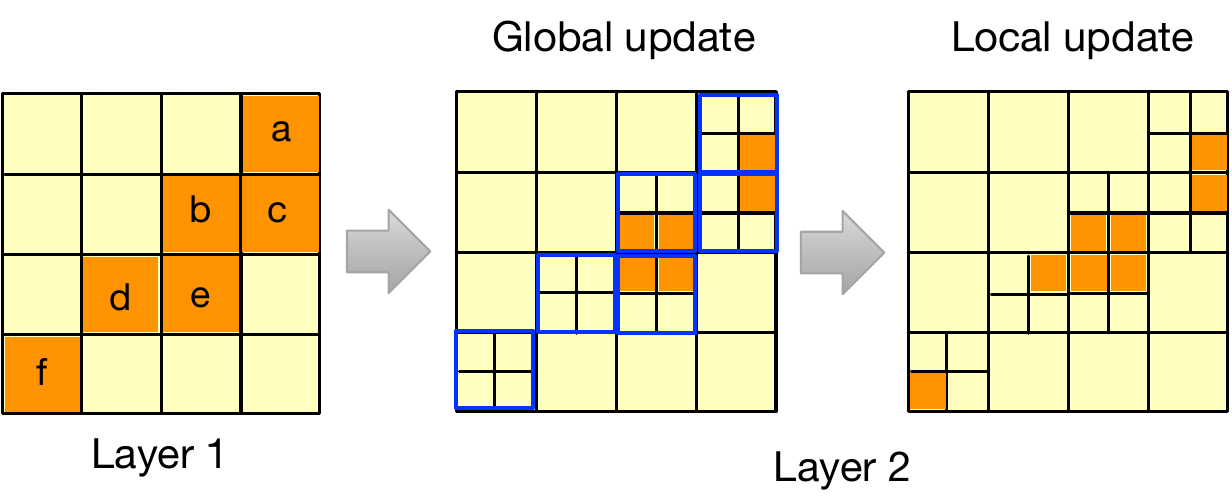}
  \caption{An illustration of the proposed two-step mesh refinement approach for critical region detection.}   
  \label{fig:twoStep_refinement}
\end{figure}

An example of the proposed algorithm is illustrated in Figure~\ref{fig:twoStep_refinement}. Starting with a partition of cells passed from layer 1, the global update selects two partitioned bins from cells $b$ and $e$, as well as one bin from cells $a$ and $c$, accounting for a quarter of the total bins. For cells $d$ and $f$, which are entirely missed in the global update, the local update selects one out of the four partitioned bins from each. It is important to note that the thresholds for the global and local updates can be different. The combination of global and local updates results in a balanced detection at both global and regional scales and helps to preserve continuous features.

The choices of mesh refinement parameters, including the bin widths ${k_l}$ for mesh partition and the thresholds ${p^{th}}$ for global and local updates, directly impact the topology of RoIs and buffer zones, and thus influence the accuracy of region detection and compression ratios. Ideally, we aim for RoIs that exhibit continuity to avoid putting excessive buffer zones along their circumference. Additionally, the choice of parameters should be adaptable to capture both nodal and areal features, as well as features in both turbulent and smooth backgrounds. Failing to capture featured regions will result in large errors in QoIs, while an oversized RoI will reduce the compression ratio as more data is compressed using smaller error bounds. 

\begin{figure}[t]
        \centering
        \includegraphics[width=0.6\linewidth]{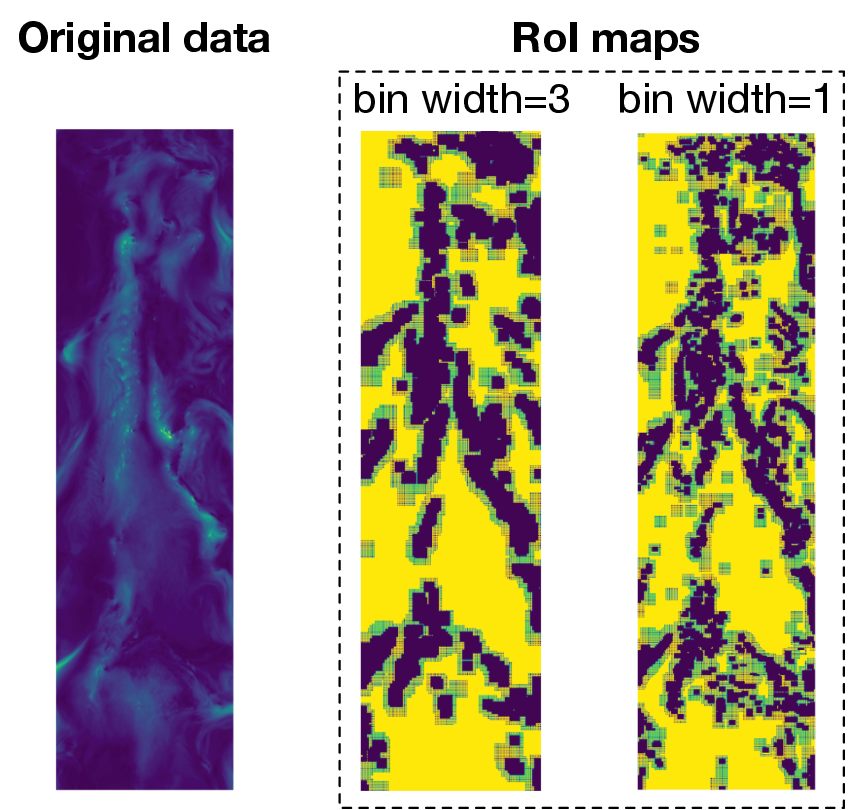}
        \caption{The impact of mesh refinement bins on the resulting topology of RoIs. The middle and right sub-figures depict the mask of acquired RoI/buffer zones.}
    \label{fig:map-binwidth}
\end{figure}

We summarize three rules for choosing mesh refinement parameters based on theoretical analyses. First, using coarser bins for mesh refinement leads to clustered RoIs, while a refined partition helps resolve fine structures. Figure~\ref{fig:map-binwidth} uses an variable of integrated vapor transport outputted from E3SM simulation as an example. 
Here, we implement a single layer of mesh refinement with different bin widths, and keep the total selected regions the same. As the figure shows, a bin width of $3\times3$ returns more continuous RoIs, however a bin width of $1\times1$ covers the filamentary structure in refined scales. Second, multiple layers help isolate features from a noisy background but tend to generate scattered RoIs when data variability is high. Comparing to clustered RoIs, the scattered regions need larger buffer zones, which in turn leads to smaller compression rates.     
Third, for multi-layered mesh refinement, using a set of thresholds that start with a high bar (i.e., fewer data survive) and then progressively lower bars helps capture distributed features, while the opposite choice is suitable for capturing features of continuous structures.

\section{Experiments}\label{sec:evaluation}
All experiments were conducted on the high-performance Andes cluster at Oak Ridge National Laboratory~\cite{Andes}. The cluster consists of nodes with two 16-core AMD EPYC 7302 processors and 256 GB of memory. The GCC 10.1 compiler was used for compilation of the codes. 
\subsection{Climate datasets \& analysis codes}
The climate datasets used for testing were obtained from a high-resolution (HR) configuration atmosphere-only E3SM simulation spanning one year. We generated data outputs at two separate temporal rates: 6 hourly and hourly. The HR configuration corresponds to a grid resolution of approximately 25 km, resulting in around 350,000 \texttt{float32} data points per variable per time-slice. The E3SM simulation utilizes a cubed-sphere computation grid representing latitude-longitude units. To enhance the compression performance, we transformed the coordinates from 1{\scshape d} longitude-latitude units to 2{\scshape d} Cartesian space. For each variable, we buffered every 120 consecutive timesteps and compressed them at once.

We utilized TempestExtreme Version 2.1~\cite{ullrich2021tempestextremes}, which is a software package for feature tracking and scientific analysis of global Earth-system data. The TC tracking algorithm in TempestExtreme first identifies candidate TC nodes in each time slice based on criteria such as minimum sea-level pressure and a coherent upper-level warm core associated with a local low-pressure area. It then applies various exclusionary criteria, including minimum wind speed, maximum travel distance, minimum life cycle, and maximum allowable gap between any two consecutive nodes in a trajectory, to isolate qualified candidates and stitch them together to form trajectories across time steps. The five variables used as inputs for TC tracking are pressure at sea-level (PSL), temperature at 500 hPa and 200 hPa (T500, T200), and zonal and meridional wind speed at the surface (UBOT, VBOT). Throughout the experiments, we adopted the same threshold settings as suggested in \cite{ullrich2021tempestextremes} to detect TCs tracks from both hourly and 6-hourly data. To avoid rounding biases, all time-relevant thresholds were multiples of 6. The TC tracks detected from the uncompressed hourly data were used as the ground truth for evaluation purposes. 

\subsection{Lossless compression}
We begin by measuring the compression ratios achieved by lossless compressors. Specifically, we selected Zstd~\cite{collet2021rfc}, a hybrid lossless compression algorithm known for its ability to provide fast compression and high reduction ratios. Zstd has a stable format and has been published as IETF RFC 8878. We applied Zstd to compress five TC tracking variables in the hourly dataset and obtained the following compression ratios (CR) for 5 variables: CR(PSL) = $1.31$, CR(T200) = $1.26$, CR(T500) = $1.28$, CR(UBOT) = $1.08$, and CR(VBOT) = $1.08$. These results illustrate the limited compression ratios that can be achieved by lossless compressors on floating-point scientific data. 

\subsection{Reduction in temporal-resolution vs reduction in precision}
\label{sec:hourly_vs_6hourly_comparison}
Next, we evaluated the impact of precision and temporal resolution on feature preservation by analyzing the trajectories and key characteristics of TC tracks obtained from datasets reduced by timestep decimation and precision reduction. For precision reduction, we employed three state-of-the-art floating-point lossy compressors: MGARD, ZFP, and SZ, to compress the hourly dataset by a factor of 6, ensuring the compressed data size matches that of the 6-hourly snapshot data. Notably, these three lossy compressors only support applying uniform error bounds across the entire data space.  

\begin{figure}[h]
    \begin{subfigure}[b]{.475\linewidth}
    	\centering
        \includegraphics[width=0.875\linewidth]{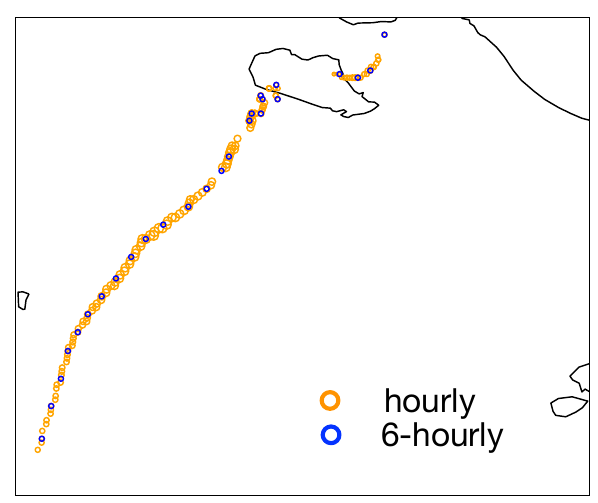}
    \caption{Generally matched TC tracks}
    \end{subfigure}\hfill
    \begin{subfigure}[b]{.475\linewidth}
    	\centering
        \includegraphics[width=0.875\linewidth]{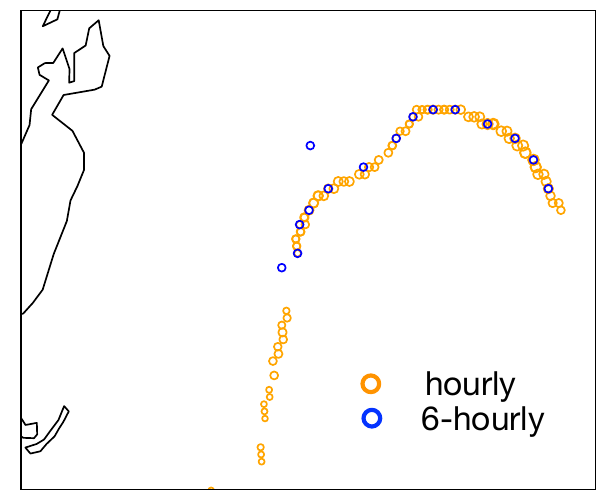}
    \caption{Partially matched TC tracks}
    \end{subfigure}
     \caption{Visualize the generally and partially matched TC tracks classified by curve-similarity.}
     \label{fig:tc_curve_distance}
\end{figure}

\begin{figure}
    \centering
    \includegraphics[width=.9\linewidth]{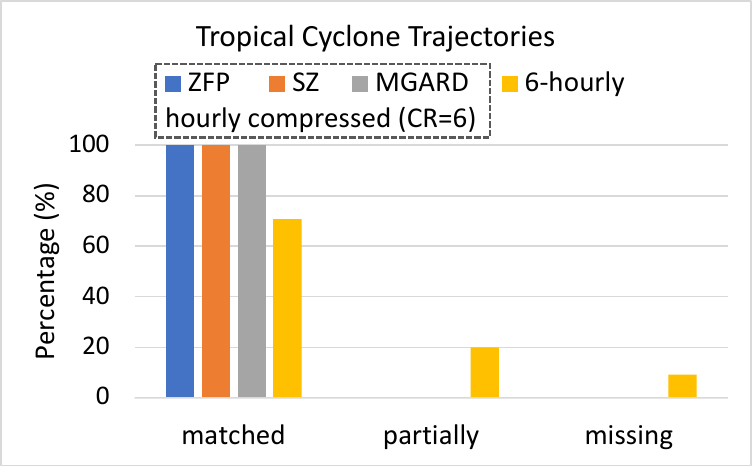}
    \caption{Errors of TC tracks measured through curve-similarity using data reduced by three uniform-error-bounded lossy compressors at a compression ratio of 6 and by timestep decimation (i.e., 6-hourly).}
    \label{fig:TC_6hourly_vs_uniform_curve}
\end{figure}

We propose to measure the disparity of TC tracks found in dataset reduced by two different approaches with those found in the hourly uncompressed dataset using curve similarity, as the method hourly dataset contains more time-slices than the 6-hourly dataset, excluding the use of a point-wise metric. We utilized two types of curve distance: discrete Frechet~\cite{eiter1994computing} and partial curve mapping (PCM)~\cite{witowski2012parameter}. Based on the scores from these two measures, we categorized the TC tracks detected from the reduced data into three groups: matched, partially matched, and missed. Figure~\ref{fig:tc_curve_distance} provides an example illustrating matched and partially matched cases classified by curve similarity.
Figure~\ref{fig:TC_6hourly_vs_uniform_curve} illustrates the classification results obtained from 6-hourly uncompressed data and hourly data compressed using uniform error bounds by MGARD, SZ, and ZFP. It shows that approximately $70.71\%$ of the TC tracks found in the 6-hourly dataset are classified as generally matched, while $9.29\%$ of them are considered as missed. In comparison, at the same storage cost (i.e., compression ratio of 6), the TC tracks generated by hourly data reduced by the three lossy compressors achieve a $100\%$ matching rate.

\begin{figure}[h]
\centering
\includegraphics[width=\linewidth]{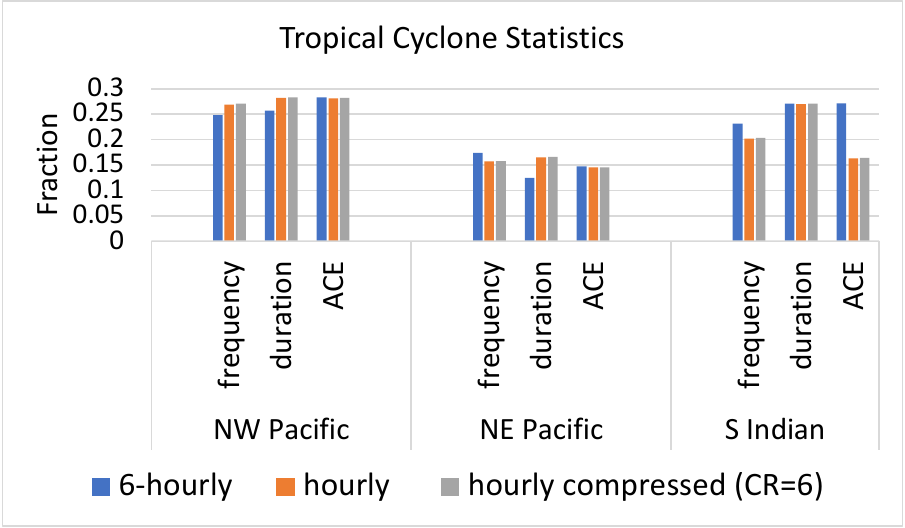}
\caption{The area-to-global TC statistical distribution measured at different geographic regions using hourly, hourly compressed (6X precision-based), and 6-hourly data.}
\label{fig:tc_statistics}
\end{figure}

The key TC characteristics commonly used for climate model evaluation and detection algorithm validation are the global TC frequency, duration of the life cycle, and accumulated cyclone energy (ACE)~\cite{zarzycki2017assessing, zarzycki2021metrics}. TC frequency represents the number of TC events in a given year, duration denotes the lifetime of TCs, and ACE is calculated by summing the square of a TC's maximum sustained wind speed every six hours. We selected three specific geographic areas - the Northwest Pacific, Northeast Pacific, and South Indian - and computed these TC characteristics. We employed the area-to-global probability distribution as the error metric, which is the ratio of TC characteristics in a focused area to the measurements of the global region.
Figure~\ref{fig:tc_statistics} presents the TC statistics measured over the span of a year obtained from hourly and 6-hourly uncompressed data and hourly data lossy compressed using MGARD by a factor of 6. Across all geographic areas and for all three TC characteristics, the maximum error between the hourly uncompressed data and data compressed by a factor of $6$ is only $0.75\%$. In comparison, the 6-hourly temporally decimated data exhibits a discrepancy of approximately $10\%$ in most cases, with a maximum error of $66.6\%$ observed for ACE measured in the South Indian region. With the same storage cost, the lossy compressed data saved in original temporal resolution better preserves the QoIs of TC variables compared to the uncompressed data saved in lower temporal resolution.

\subsection{Region-adaptive compression in spatiotemporal domain}
We conducted further evaluations to assess the advantages of using multiple error-bounded, region-adaptive compression techniques in preserving QoIs, and compared the results against those obtained using uniformly error-bounded compressors. For the adaptive compression, we performed critical region detection only on PSL, as the thresholds for TC detection are initially defined based on its vortex intensity. We then used the detected RoI maps to compress the rest 4 variables. To closely examine the errors introduced by lossy compression, we compared the cyclone locations from the compressed and uncompressed hourly data by computing the point-wise great-circle distance (gcd) using the following equation~\cite{ullrich2021tempestextremes}:
\begin{equation*}
\textrm{gcd}(\lambda,\varphi;\tilde{\lambda},\tilde{\varphi}) = \arccos{(\sin{\varphi}\sin{\tilde{\varphi}} + \cos{\varphi}\cos{\tilde{\varphi}}\cos{(\lambda-\tilde{\lambda})})},
\end{equation*}
where $\{\lambda, \varphi\}$ and $\{\tilde{\lambda},\tilde{\varphi}\}$ represent the latitude-longitude coordinates of a cyclone in the original and reduced datasets, respectively. We averaged the gcd of all TC locations along a path to measure the error of a TC track.

\begin{figure*}[h]
\centering
\begin{subfigure}[b]{0.32\linewidth}
    \centering
\includegraphics[width=\textwidth]{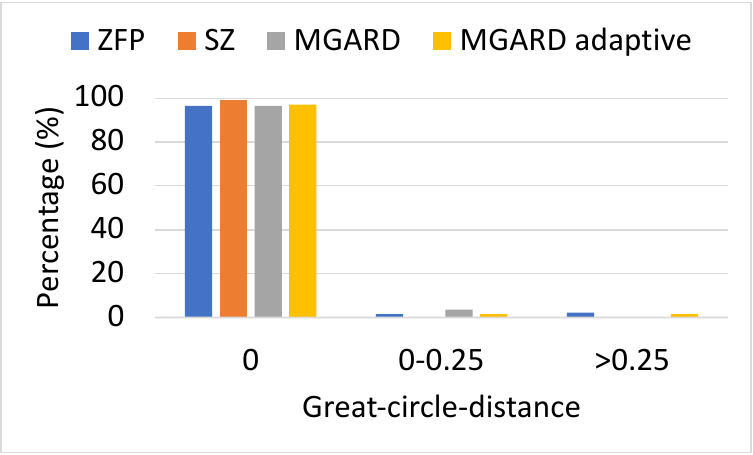}
\caption{Compression ratio = 6}
\label{fig:TC_curve_err_vs_cr}
\end{subfigure}
\hfill
\begin{subfigure}[b]{0.32\linewidth}
\centering
\includegraphics[width=\linewidth]{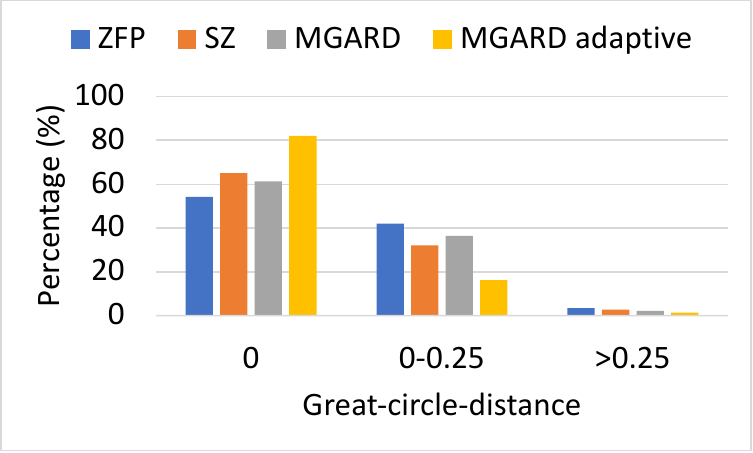}
\caption{Compression ratio = 13}
\label{fig:TC_gcd_err_vs_cr}
\end{subfigure}
\hfill
\begin{subfigure}[b]{0.32\linewidth}
\centering
\includegraphics[width=\linewidth]{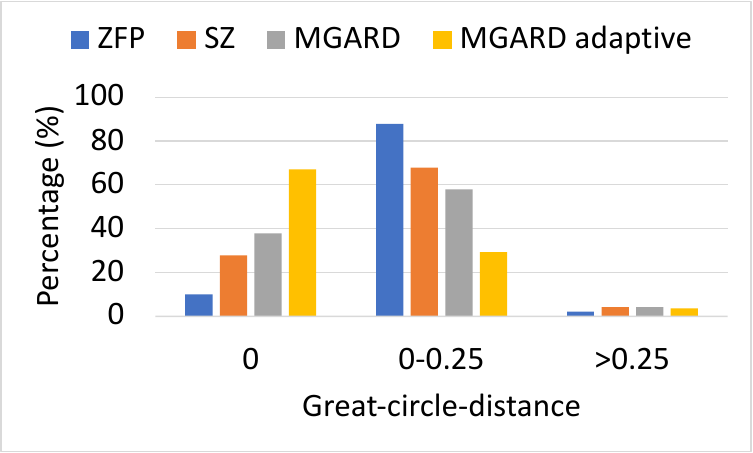}
\caption{Compression ratio = 23}
\label{fig:TC_gcd_err_vs_cr}
\end{subfigure}
\caption{Errors of TC tracks measured by point-wise distance using data lossy compressed with different compressors.}
\label{fig:TC_adaptive_vs_uniform_gcd}
\end{figure*}

\begin{figure*}[h]
\centering
\begin{subfigure}[b]{0.49\linewidth}
    \centering
\includegraphics[width=\textwidth]{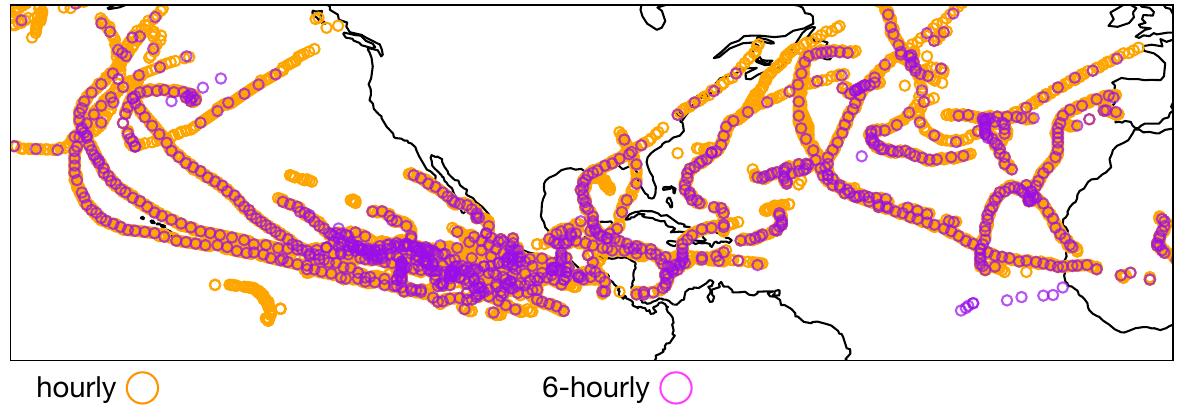}
\caption{Visualize the TC tracks found in hourly and 6-hourly data (temporal decimation rate = 6).}
\label{fig:TC_vis_hourly_vs_6hourly}
\end{subfigure}
\hfill
\begin{subfigure}[b]{0.49\linewidth}
\centering
\includegraphics[width=\linewidth]{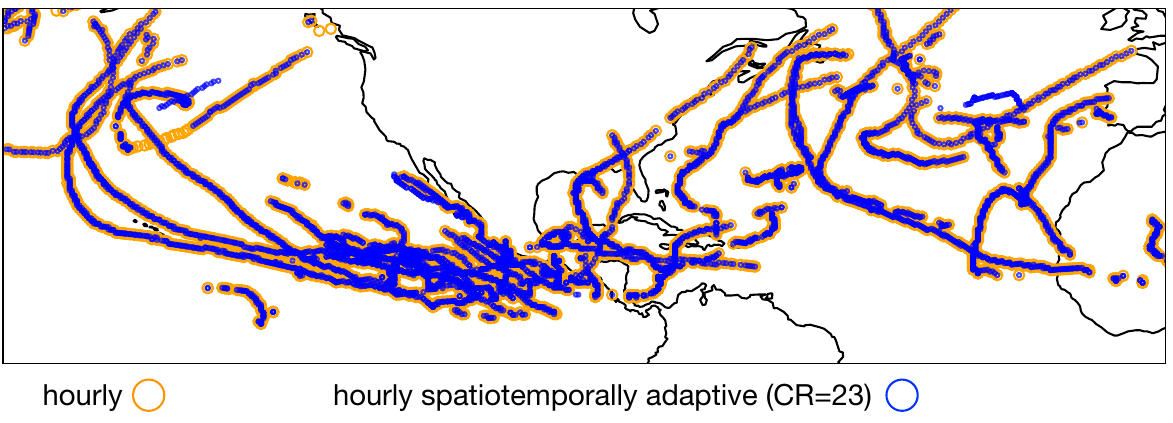}
\caption{Visualize the TC tracks found in hourly and hourly spatiotemporally compressed data (compression ratio = 23).}
\label{fig:TC_vis_hourly_vs_adaptive}
\end{subfigure}
\caption{Global distributing of TC tracks detected in hourly, 6-hourly, and spatiotemporally adaptive reduced hourly data over one year time span.}
\label{fig:TC_qualitative_eval}
\end{figure*}


Figure~\ref{fig:TC_adaptive_vs_uniform_gcd} illustrates the errors measured by the point-wise metrics when the input variables are compressed at compression ratios of 6, 13, and 23. We categorized the errors into three ranges: $0$, $[0,~0.25^\circ]$, and $[0.25^\circ,~\infty]$, where $0.25^\circ$ represents the physical grid spacing of the HR simulation configuration. The percentage of TC tracks with an averaged gcd falling into each error range is plotted. The results demonstrate that the advantage of using spatiotemporally adaptive compression becomes more prominent as larger compression ratios are employed. At a compression ratio of 6, the uniform and spatiotemporal adaptive compressors have all obtained very accurate results, leaving limited room for improvement with the adaptive approach. Notably, the adaptive compression shows slightly larger errors due to mismatches between the compressor-detected RoIs and the actual locations of TC vortexes. At a compression ratio of 13, the adaptive method yields $26.37\%$ to $51.31\%$ more perfectly matched TC tracks (i.e., gcd=0) compared to uniform compressors, and $48.88\%$ to $61.01\%$ fewer TC tracks in the $[0,~0.25]$ and $[0.25,~\infty]$ error ranges. At a compression ratio of 23, the percentage of TC tracks with errors in the $[0.25,~\infty]$ range is similar among the three uniform and adaptive compressors. However, the adaptive compressor achieves $77.34\%$ to $571.4\%$ more perfectly matched TC tracks and $49.37\%$ to $66.66\%$ fewer TC tracks in the $[0,~0.25]$ error range compared to the uniform error-bounded compressors. Among the three uniform compressors, data compressed using ZFP exhibits the least number of perfectly matched TC locations due to its filtering of high-frequency components, which correspond to vortex features of cyclones.

\begin{figure}
    \centering
    \includegraphics[width=.95\linewidth]{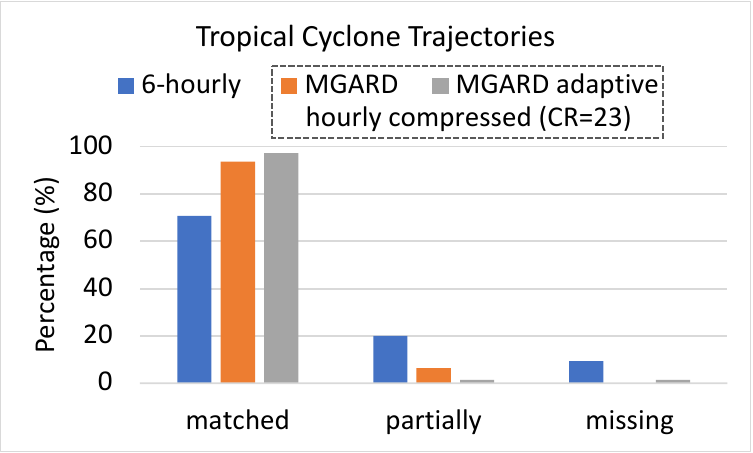}
    \caption{Errors of TC tracks measured through curve-similarity using hourly data reduced by MGARD adaptive compression, MGARD uniform compression at a ratio of 23, and 6-hourly uncompressed data.}
    \label{fig:TC_adaptive_vs_uniform_curve}
\end{figure}

Finally, we visualize the TC tracks obtained from 6-hourly uncompressed data and hourly data reduced using the proposed spatiotemporal adaptive compressor by a factor of 23 in Figure~\ref{fig:TC_qualitative_eval}. The TC tracks derived from adaptive compression exhibit significantly better alignment with the uncompressed results compared to those obtained through temporal decimation, while requiring only a quarter of storage.  Figure~\ref{fig:TC_adaptive_vs_uniform_curve} employs the curve-similarity metric proposed in Section~\ref{sec:hourly_vs_6hourly_comparison} to provide a quantitative comparison of the TC tracks displayed in Figure~\ref{fig:TC_qualitative_eval}, as well as those reduced using uniform error-bounded compression. The results confirm that the TC tracks obtained from spatiotemporal adaptive compressed data are the most accurate and have fewer missing paths.

\subsection{Throughput overhead}
It is crucial to ensure that the adaptive error-bounded compression does not incur significant computational overhead compared to uniform compressors. To achieve this, we leverage the multilevel decomposed coefficients, which are intermediate products during MGARD compression, for RoI detection, reducing the cost of critical region detection. The reconstruction of adaptive compressed data can be performed using the same routine as data reduced by MGARD, with the application of a specialized encoding method discussed in~\cite{gong2022region}.

When compared to single-error-bounded MGARD, the computational overhead of adaptive compression primarily arises from the mesh refinement process used for critical region detection and the construction of the buffer zone. In our evaluation, we focused on compressing the TC tracking variable in a 3 {\scshape d} spatiotemporal space and measured the overhead with different mesh refinement parameters. Each measurement was performed 50 times, and the average number was reported.

\begin{figure}[t]
    \centering
    \includegraphics[width=0.85\linewidth]{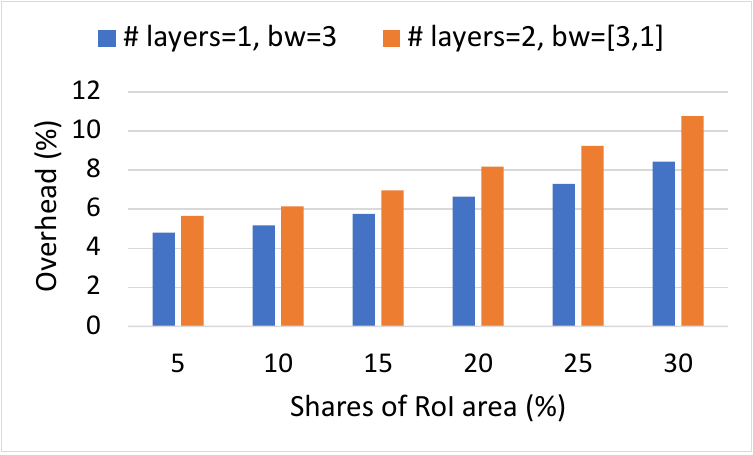}
    \caption{Overhead of region-adaptive compression in spatial-temporal domain and the relation to two mesh refinement parameters -- layers, bin width (bw).}
    \label{fig:overhead}
\end{figure}

Figure~\ref{fig:overhead} illustrates the ratio of the spatiotemporally adaptive compression overhead to the rest of the computation. Generally, the overhead of adaptive compression is proportional to the size of the RoIs. As discussed in Section~\ref{sec:adaptive-region-detection}, users have the flexibility to adjust the topology of RoIs by varying the number of layers and the bin width (bw) used in mesh refinement. The figure demonstrates that utilizing multiple layers in mesh refinement increases the computational overhead due to the additional cost associated with mesh refinement and buffer zone construction. However, within the tested range of $5\%$ to $30\%$, the overhead of spatiotemporally adaptive compression remained at a modest level of only $5\%$ to $11\%$.

\section{Conclusion and Future Work}\label{sec:conclusion}
In this paper, we have conducted a comprehensive assessment to understand the impact of data reduction on spatiotemporal analyses using temporal decimation and error-bounded lossy compression techniques. Using the simulation data generated by a well-known climate model, E3SM, we have demonstrated that compressing spatiotemporal data at higher frequencies, as opposed to traditional temporal decimation techniques, can lead to significant reductions in storage size while simultaneously improving the quality of analysis. This finding highlights the potential of leveraging lossy compression methods for scientific data analysis. 

To address the specific requirements of post-analyses that rely on tracking variational features across timestep data, we have introduced a novel spatiotemporal adaptive lossy compression framework. This framework incorporates an optimized critical region detection algorithm and error preservation theories in spatiotemporal space. We have also provided parameter choices for identifying features of different structures and typologies. 
Our evaluation has shown that the data reduced by this adaptive compression framework outperforms three state-of-the-art lossy compressors in terms of accuracy in TC tracking results, with a merely $5\%$ to $11\%$ computational overhead. 

The accuracy of RoI detection emerges as a critical factor influencing the performance of adaptive compression. Looking ahead, our future work will focus on combining various RoI detection methods, such as machine learning and statistical sampling, with the proposed non-uniform error bounded compression framework. This integration has the potential to further enhance the capabilities and efficiency of our approach.




\bibliographystyle{IEEEtran}
\bibliography{IEEEexample}

\begin{thebibliography}{10}
\providecommand{\url}[1]{#1}
\csname url@samestyle\endcsname
\providecommand{\newblock}{\relax}
\providecommand{\bibinfo}[2]{#2}
\providecommand{\BIBentrySTDinterwordspacing}{\spaceskip=0pt\relax}
\providecommand{\BIBentryALTinterwordstretchfactor}{4}
\providecommand{\BIBentryALTinterwordspacing}{\spaceskip=\fontdimen2\font plus
\BIBentryALTinterwordstretchfactor\fontdimen3\font minus
  \fontdimen4\font\relax}
\providecommand{\BIBforeignlanguage}[2]{{%
\expandafter\ifx\csname l@#1\endcsname\relax
\typeout{** WARNING: IEEEtran.bst: No hyphenation pattern has been}%
\typeout{** loaded for the language `#1'. Using the pattern for}%
\typeout{** the default language instead.}%
\else
\language=\csname l@#1\endcsname
\fi
#2}}
\providecommand{\BIBdecl}{\relax}
\BIBdecl

\bibitem{frontier}
\BIBentryALTinterwordspacing
Frontier exscale supercomputer. [Online]. Available:
  \url{https://www.olcf.ornl.gov/frontier}
\BIBentrySTDinterwordspacing

\bibitem{summit}
\BIBentryALTinterwordspacing
Summit supercomputer:. [Online]. Available:
  \url{https://www.olcf.ornl.gov/summit}
\BIBentrySTDinterwordspacing

\bibitem{chang2004retrieving}
C.-C. Chang, J.-C. Chuang, and Y.-S. Hu, ``Retrieving digital images from a
  jpeg compressed image database,'' \emph{Image and Vision Computing}, vol.~22,
  no.~6, pp. 471--484, 2004.

\bibitem{deri2012tsdb}
L.~Deri, S.~Mainardi, and F.~Fusco, ``tsdb: A compressed database for time
  series,'' in \emph{International Workshop on Traffic Monitoring and
  Analysis}.\hskip 1em plus 0.5em minus 0.4em\relax Springer, 2012, pp.
  143--156.

\bibitem{chen2001query}
Z.~Chen, J.~Gehrke, and F.~Korn, ``Query optimization in compressed database
  systems,'' in \emph{Proceedings of the 2001 ACM SIGMOD international
  conference on Management of data}, 2001, pp. 271--282.

\bibitem{arion2007xquec}
A.~Arion, A.~Bonifati, I.~Manolescu, and A.~Pugliese, ``Xquec: A
  query-conscious compressed xml database,'' \emph{ACM Transactions on Internet
  Technology (TOIT)}, vol.~7, no.~2, pp. 10--es, 2007.

\bibitem{zarzycki2017assessing}
C.~M. Zarzycki and P.~A. Ullrich, ``Assessing sensitivities in algorithmic
  detection of tropical cyclones in climate data,'' \emph{Geophysical Research
  Letters}, vol.~44, no.~2, pp. 1141--1149, 2017.

\bibitem{zarzycki2021metrics}
C.~M. Zarzycki, P.~A. Ullrich, and K.~A. Reed, ``Metrics for evaluating
  tropical cyclones in climate data,'' \emph{Journal of Applied Meteorology and
  Climatology}, vol.~60, no.~5, pp. 643--660, 2021.

\bibitem{mcclenny2020sensitivity}
E.~E. McClenny, P.~A. Ullrich, and R.~Grotjahn, ``Sensitivity of atmospheric
  river vapor transport and precipitation to uniform sea surface temperature
  increases,'' \emph{Journal of Geophysical Research: Atmospheres}, vol. 125,
  no.~21, p. e2020JD033421, 2020.

\bibitem{zhou2021uncertainties}
Y.~Zhou, T.~A. O'Brien, P.~A. Ullrich, W.~D. Collins, C.~M. Patricola, and
  A.~M. Rhoades, ``Uncertainties in atmospheric river lifecycles by detection
  algorithms: climatology and variability,'' \emph{Journal of Geophysical
  Research: Atmospheres}, vol. 126, no.~8, p. e2020JD033711, 2021.

\bibitem{tao2017significantly}
D.~Tao, S.~Di, Z.~Chen, and F.~Cappello, ``Significantly improving lossy
  compression for scientific data sets based on multidimensional prediction and
  error-controlled quantization,'' in \emph{2017 IEEE International Parallel
  and Distributed Processing Symposium (IPDPS)}.\hskip 1em plus 0.5em minus
  0.4em\relax IEEE, 2017, pp. 1129--1139.

\bibitem{liang2018error}
X.~Liang, S.~Di, D.~Tao, S.~Li, S.~Li, H.~Guo, Z.~Chen, and F.~Cappello,
  ``Error-controlled lossy compression optimized for high compression ratios of
  scientific datasets,'' in \emph{2018 IEEE International Conference on Big
  Data (Big Data)}.\hskip 1em plus 0.5em minus 0.4em\relax IEEE, 2018, pp.
  438--447.

\bibitem{zhao2021optimizing}
K.~Zhao, S.~Di, M.~Dmitriev, T.-L.~D. Tonellot, Z.~Chen, and F.~Cappello,
  ``Optimizing error-bounded lossy compression for scientific data by dynamic
  spline interpolation,'' in \emph{2021 IEEE 37th International Conference on
  Data Engineering (ICDE)}.\hskip 1em plus 0.5em minus 0.4em\relax IEEE, 2021,
  pp. 1643--1654.

\bibitem{lindstrom2006fast}
P.~Lindstrom and M.~Isenburg, ``Fast and efficient compression of
  floating-point data,'' \emph{IEEE transactions on visualization and computer
  graphics}, vol.~12, no.~5, pp. 1245--1250, 2006.

\bibitem{lindstrom2014fixed}
P.~Lindstrom, ``Fixed-rate compressed floating-point arrays,'' \emph{IEEE
  transactions on visualization and computer graphics}, vol.~20, no.~12, pp.
  2674--2683, 2014.

\bibitem{ainsworth2017compression}
M.~Ainsworth, S.~Klasky, and B.~Whitney, ``Compression using lossless
  decimation: analysis and application,'' \emph{SIAM Journal on Scientific
  Computing}, vol.~39, no.~4, pp. B732--B757, 2017.

\bibitem{ainsworth2019multilevel}
M.~Ainsworth, O.~Tugluk, B.~Whitney, and S.~Klasky, ``Multilevel techniques for
  compression and reduction of scientific data---the multivariate case,''
  \emph{SIAM Journal on Scientific Computing}, vol.~41, no.~2, pp.
  A1278--A1303, 2019.

\bibitem{ainsworth2019qoi}
------, ``Multilevel techniques for compression and reduction of scientific
  data--quantitative control of accuracy in derived quantities,'' \emph{SIAM
  Journal on Scientific Computing}, vol.~41, no.~4, pp. A2146--A2171, 2019.

\bibitem{jiao2022toward}
P.~Jiao, S.~Di, H.~Guo, K.~Zhao, J.~Tian, D.~Tao, X.~Liang, and F.~Cappello,
  ``Toward quantity-of-interest preserving lossy compression for scientific
  data,'' \emph{Proceedings of the VLDB Endowment}, vol.~16, no.~4, pp.
  697--710, 2022.

\bibitem{liang2020toward}
X.~Liang, H.~Guo, S.~Di, F.~Cappello, M.~Raj, C.~Liu, K.~Ono, Z.~Chen, and
  T.~Peterka, ``Toward feature-preserving 2d and 3d vector field compression.''
  in \emph{PacificVis}, 2020, pp. 81--90.

\bibitem{gong2022region}
Q.~Gong, B.~Whitney, C.~Zhang, X.~Liang, A.~Rangarajan, J.~Chen, L.~Wan,
  P.~Ullrich, Q.~Liu, R.~Jacob \emph{et~al.}, ``Region-adaptive,
  error-controlled scientific data compression using multilevel
  decomposition,'' in \emph{Proceedings of the 34th International Conference on
  Scientific and Statistical Database Management}, 2022, pp. 1--12.

\bibitem{liang2022toward}
X.~Liang, S.~Di, F.~Cappello, M.~Raj, C.~Liu, K.~Ono, Z.~Chen, T.~Peterka, and
  H.~Guo, ``Toward feature-preserving vector field compression,'' \emph{IEEE
  Transactions on Visualization and Computer Graphics}, 2022.

\bibitem{caldwell2019doe}
P.~M. Caldwell, A.~Mametjanov, Q.~Tang, L.~P. Van~Roekel, J.-C. Golaz, W.~Lin,
  D.~C. Bader, N.~D. Keen, Y.~Feng, R.~Jacob \emph{et~al.}, ``The doe e3sm
  coupled model version 1: Description and results at high resolution,''
  \emph{Journal of Advances in Modeling Earth Systems}, vol.~11, no.~12, pp.
  4095--4146, 2019.

\bibitem{mendelsohn2012impact}
R.~Mendelsohn, K.~Emanuel, S.~Chonabayashi, and L.~Bakkensen, ``The impact of
  climate change on global tropical cyclone damage,'' \emph{Nature climate
  change}, vol.~2, no.~3, pp. 205--209, 2012.

\bibitem{deutsch1996gzip}
P.~Deutsch \emph{et~al.}, ``Gzip file format specification version 4.3,'' 1996.

\bibitem{burtscher2008fpc}
M.~Burtscher and P.~Ratanaworabhan, ``Fpc: A high-speed compressor for
  double-precision floating-point data,'' \emph{IEEE Transactions on
  Computers}, vol.~58, no.~1, pp. 18--31, 2008.

\bibitem{collet2021rfc}
Y.~Collet, ``Rfc 8878: Zstandard compression and the'application/zstd'media
  type,'' 2021.

\bibitem{lakshminarasimhan2013isabela}
S.~Lakshminarasimhan, N.~Shah, S.~Ethier, S.-H. Ku, C.-S. Chang, S.~Klasky,
  R.~Latham, R.~Ross, and N.~F. Samatova, ``Isabela for effective in situ
  compression of scientific data,'' \emph{Concurrency and Computation: Practice
  and Experience}, vol.~25, no.~4, pp. 524--540, 2013.

\bibitem{ainsworth2018multilevel}
M.~Ainsworth, O.~Tugluk, B.~Whitney, and S.~Klasky, ``Multilevel techniques for
  compression and reduction of scientific data--the univariate case,''
  \emph{Computing and Visualization in Science}, vol.~19, no.~5, pp. 65--76,
  2018.

\bibitem{lee2022error}
J.~Lee, Q.~Gong, J.~Choi, T.~Banerjee, S.~Klasky, S.~Ranka, and A.~Rangarajan,
  ``Error-bounded learned scientific data compression with preservation of
  derived quantities,'' \emph{Applied Sciences}, vol.~12, no.~13, p. 6718,
  2022.

\bibitem{gong2021maintaining}
Q.~Gong, X.~Liang, B.~Whitney, J.~Y. Choi, J.~Chen, L.~Wan, S.~Ethier, S.-H.
  Ku, R.~M. Churchill, C.-S. Chang \emph{et~al.}, ``Maintaining trust in
  reduction: Preserving the accuracy of quantities of interest for lossy
  compression,'' in \emph{Smoky Mountains Computational Sciences and
  Engineering Conference}.\hskip 1em plus 0.5em minus 0.4em\relax Springer,
  2021, pp. 22--39.

\bibitem{banerjee2022algorithmic}
T.~Banerjee, J.~Choi, J.~Lee, Q.~Gong, R.~Wang, S.~Klasky, A.~Rangarajan, and
  S.~Ranka, ``An algorithmic and software pipeline for very large scale
  scientific data compression with error guarantees,'' in \emph{2022 IEEE 29th
  International Conference on High Performance Computing, Data, and Analytics
  (HiPC)}.\hskip 1em plus 0.5em minus 0.4em\relax IEEE, 2022, pp. 226--235.

\bibitem{moffat2019huffman}
A.~Moffat, ``Huffman coding,'' \emph{ACM Computing Surveys (CSUR)}, vol.~52,
  no.~4, pp. 1--35, 2019.

\bibitem{cappello2019use}
F.~Cappello, S.~Di, S.~Li, X.~Liang, A.~M. Gok, D.~Tao, C.~H. Yoon, X.-C. Wu,
  Y.~Alexeev, and F.~T. Chong, ``Use cases of lossy compression for
  floating-point data in scientific data sets,'' \emph{The International
  Journal of High Performance Computing Applications}, vol.~33, no.~6, pp.
  1201--1220, 2019.

\bibitem{diffenderfer2019error}
J.~Diffenderfer, A.~L. Fox, J.~A. Hittinger, G.~Sanders, and P.~G. Lindstrom,
  ``Error analysis of zfp compression for floating-point data,'' \emph{SIAM
  Journal on Scientific Computing}, vol.~41, no.~3, pp. A1867--A1898, 2019.

\bibitem{berg2018precipitation}
P.~Berg, O.~Christensen, K.~Klehmet, G.~Lenderink, J.~Olsson, C.~Teichmann, and
  W.~Yang, ``Precipitation extremes in a euro-cordex 0.11° ensemble at hourly
  resolution,'' \emph{Nat. Hazards Earth Syst. Sci}, pp. 1--21, 2018.

\bibitem{meyer2021more}
J.~Meyer, M.~Neuper, L.~Mathias, E.~Zehe, and L.~Pfister, ``More frequent flash
  flood events and extreme precipitation favouring atmospheric conditions in
  temperate regions of europe,'' \emph{Hydrology and Earth System Sciences
  Discussions}, vol. 2021, pp. 1--28, 2021.

\bibitem{Andes}
\BIBentryALTinterwordspacing
Andes cluster. [Online]. Available:
  \url{https://www.olcf.ornl.gov/olcf-resources/compute-systems/andes/}
\BIBentrySTDinterwordspacing

\bibitem{ullrich2021tempestextremes}
P.~A. Ullrich, C.~M. Zarzycki, E.~E. McClenny, M.~C. Pinheiro, A.~M.
  Stansfield, and K.~A. Reed, ``Tempestextremes v2. 1: a community framework
  for feature detection, tracking, and analysis in large datasets,''
  \emph{Geoscientific Model Development}, vol.~14, no.~8, pp. 5023--5048, 2021.

\bibitem{eiter1994computing}
T.~Eiter and H.~Mannila, ``Computing discrete fr{\'e}chet distance,'' 1994.

\bibitem{witowski2012parameter}
K.~Witowski and N.~Stander, ``Parameter identification of hysteretic models
  using partial curve mapping,'' in \emph{12th AIAA Aviation Technology,
  Integration, and Operations (ATIO) Conference and 14th AIAA/ISSMO
  Multidisciplinary Analysis and Optimization Conference}, 2012, p. 5580.

\end{thebibliography}

\end{document}